\documentclass[11pt]{article}

\usepackage[review]{acl}

\usepackage{times}
\usepackage{latexsym}

\usepackage[T1]{fontenc}

\usepackage[utf8]{inputenc}

\usepackage{microtype}

\usepackage{inconsolata}

\usepackage{graphicx}
\author{Kesen Wang \\
 Humain \\
  \texttt{kwang@humain.ai} \\\And
  Daulet Toibazar \\
  Humain  \\
  \texttt{dtoibazar@humain.ai}
  \\\And
 Pedro J. Moreno \\
  Humain\\
  \texttt{pmoreno@humain.ai} \\}

\title{  \includegraphics[width=0.6\textwidth]{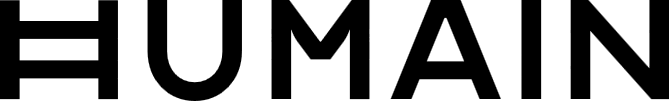} \\ \vspace{1em} $\textbf{A-SEA}^3\textbf{L}$-QA: A Fully Automated Self-Evolving, Adversarial Workflow for Arabic Long-Context Question-Answer  Generation}

\usepackage{enumitem}
\usepackage{dirtytalk}
\usepackage{booktabs}
\begin{document}
\maketitle
\begin{abstract}
We present an end-to-end, self-evolving adversarial workflow for long-context Question-Answer (QA) Generation in Arabic. By orchestrating multiple specialized LVLMs: a question generator, an evaluator, and a swarm of answer generators, our system iteratively refines its own performance without any human intervention. Starting from raw, multi-page Arabic documents across diverse domains, the question generator produces fine-grained, context-aware queries to be tackled by the answer generator swarm, and the evaluator assesses and feeds back quality metrics. This closed-loop cycle enables continuous learning: low-confidence outputs trigger automated re-generation and model updates, progressively enhancing question difficulty and relevance. Moreover, we set the quality metrics as a tunable hyperparameter, enabling question generation at controllable and customizable difficulty levels. We release \emph{\textbf{AraLongBench}}, a large-scale Arabic benchmark of single- and multi-page challenges spanning hundreds of pages, and demonstrate that our self-evolving workflow substantially outperform static pipelines, markedly boosting the long-context comprehension capabilities of leading Arabic Large Vision Language Models (LVLMs). Lastly, we also meticulously architect a fully automated agentic workflow for long-context Arabic document collection. \footnote{\url{https://github.com/wangk0b/Self_Improving_ARA_LONG_Doc.git}}
\end{abstract}

\section{Introduction}

\label{sec:intro}
Document understanding (DU) in vision-language research remains an essential yet challenging issue, particularly for documents with complex layouts and lengthy contextual dependencies.  Over the past few years, large vision-language models (LVLMs) have achieved remarkable progress on short-context tasks involving documents. Closed-source LVLMs such as OpenAI's \emph{GPT} series \citep{achiam2023gpt, openai2024o1, openai2024o3}, Google's \emph{Gemini} \citep{gemini2024}, and Anthropic's \emph{Claude} series \citep{anthropic2024claude3}, and open-source models such as \emph{InternLM-XC2-4KHD} \citep{dong2024internlm}, \emph{LLaVA-NeXT} \citep{li2024llava}, and \emph{CogVLM} \citep{wang2023cogvlm} have all demonstrated strong performance in comprehension of documents with complex layouts when there is limited context length. The models excel on single-page visual question-answering and reasoning benchmarks, such as \emph{DocVQA}, \emph{ChartQA}, and \emph{InfographicVQA}, as well as other associated datasets \citep{mathew2021docvqa, masry2022chartqa, mathew2022infographicvqa, zhu2022towards}. This achievement showcases the promise of LVLMs for DU tasks when there is a limited context length.

However, current LVLMs struggle to generalize their success to long-context DU tasks involving multi-page documents and long-range reasoning \citep{xu2023fine}. On challenging multi-page question-answering benchmarks (e.g. \emph{MMLongBench}, \emph{LongDocURL}, \emph{M-LongDoc}), even the best LVLMs reach only about 40\% accuracy, and many perform worse than text-only LLM baselines that rely on OCR-extracted text \citep{ ma2024mmlongbench, deng2024longdocurl, chia2024m}. This shortfall highlights the difficulty LVLMs have in capturing long-range and cross-page dependencies. A primary reason is the lack of training data with diverse, fine-grained questions whose answers are distributed across multiple pages. This data scarcity is even more pronounced for low-resource languages like 
Arabic. 

Up until now, the primary Arabic DU benchmark, \emph{Camel} \citep{ghaboura2024camel} and \emph{KITAB} \citep{heakl2025kitab}, focuses on single-page question answering over short passages and reports sub-optimal accuracy for state-of-the-art models, highlighting both the scarcity of fine-grained Arabic QA data and the high error rate of existing pipelines. These limitations prevent LVLMs from capturing long-range dependencies or cross-page semantics in Arabic documents. To overcome these gaps, we propose a self-evolving, multi-LVLM collaborative workflow: autonomous layout-parsing, question-generation, and evaluation workflow that iteratively enhance knowledge depth, enrich question diversity, and refine Arabic long-document QA without human intervention, culminating in a large-scale, multi-page Arabic QA generation pipeline. 
\begin{figure}[h]
    \centering
    \includegraphics[width=0.8\linewidth]{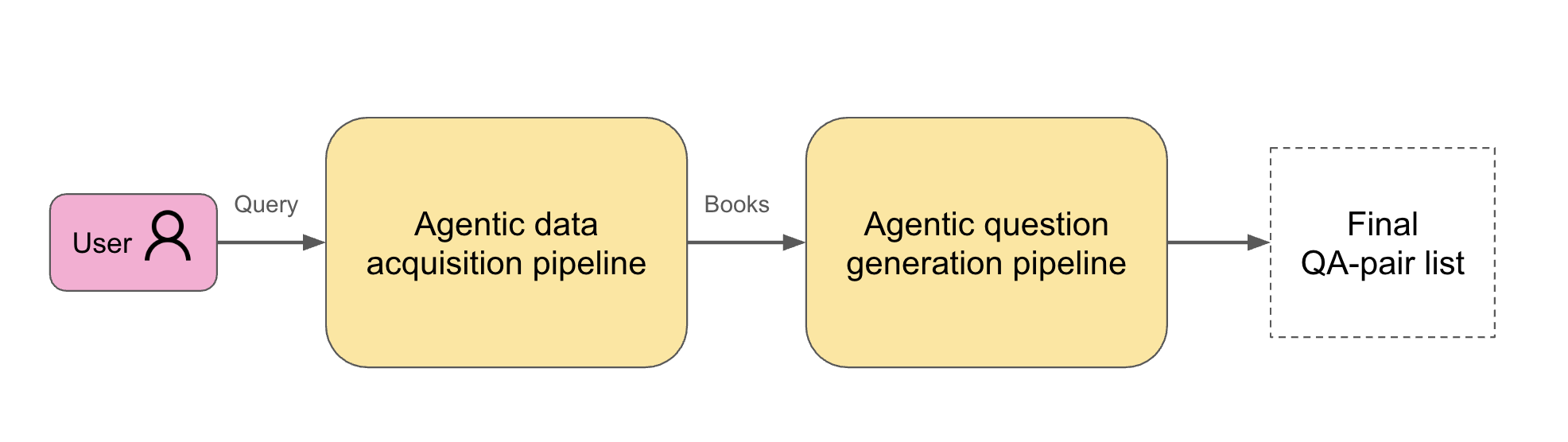}
    \caption{High-level Abstract of the Automated Pipeline}
    \label{fig:overall}
\end{figure}
\noindent In summary, our key contributions are as follows:
\begin{enumerate}[label=\arabic\*. , leftmargin=0.3in]
\item[1)] Addressing Arabic’s low-resource challenges: We design and deploy an autonomous data-collection agent to aggregate extensive and long-context Arabic corpora. 
\item[2)] Fully automated, self-evolving adversarial question generation for Arabic documents: we propose a closed-loop, automated workflow comprising layout parsing, question generation, and quality evaluation LVLMs that iteratively refine their outputs to produce high-quality multi-page Arabic QA pairs across diverse domains from long, raw documents with only a single prompt.
\item[3)] Rigorous evaluation on Arabic LVLMs: We curate \emph{\textbf{AraLongBench}}, a large-scale, multi-page Arabic QA benchmark, and perform extensive zero-shot evaluations with leading Arabic LVLMs. Results show that our generated data significantly exposes persistent weaknesses in the major LVLMs when it comes to Arabic long-context DU, guiding future model improvements.
\end{enumerate}

 Collectively, these contributions advance long-context Arabic DU by delivering an end-to-end, self-evolving  adversarial  workflow  for data annotation (Figure~\ref{fig:overall} presents a high-level abstract of the entire workflow), a publicly available benchmark, and a fully automated Arabic data acquisition pipeline, laying the groundwork for the training of more robust real-world LVLMs in long-context Arabic DU.

\section{Related Work}
\label{sec:formatting}
\subsection{Arabic DU Datasets} \label{R_BD}
A number of datasets have been developed to facilitate document understanding for various tasks, and a growing body of work has begun to address Arabic documents. There are also early Arabic layout analysis benchmarks like \emph{BCE-Arabic-v1} \citep{saad2016bce}, which brings together 1,833 scanned pages of 180 books with various fonts, multi-column layouts, photos, tables, and charts, as a benchmark for DLA, OCR, and text-to-speech research; and \emph{BADAM} \citep{kiessling2019badam}, a 400-annotated manuscript image dataset spanning historical and contemporary domains, to serve as a baseline detection benchmark in Arabic-script documents. More recent efforts have produced larger and more diverse sets. For instance, \emph{SARD} \citep{nacar2025sard} offers 843,622 synthetically created book-like images in ten Arabic fonts to offer typographic coverage and clean layouts, while \emph{KITAB-Bench} \citep{heakl2025kitab} is made up of 8,809 real-world instances in nine domains and 36 sub-domains (including tables, charts, and mixed handwritten/printed text) to evaluate modern OCR and DU methods.

Despite these advances, existing Arabic datasets remain largely restricted to single pages (scanned or artificial) and limited domains, which limits their ability to test models on long-context tasks such as cross-page co-reference, layout changes, and heavily interleaved content. To bridge this gap and enable strict testing and training of multilingual LVLMs on truly long-document Arabic material, we must develop a large-scale, multi-page Arabic DU benchmark that combines real-world diversity (books, reports, manuals, and web archives), fine-grained annotations for layout elements, tables, figures, and cross-page structures,  and automatically generated tasks covering summarization, information extraction, VQA, and reasoning. Such a dataset would open the door to the next generation of Arabic-capable LVLMs and genuinely end-to-end long-context document understanding.

\subsection{Vision-LLMs}

DU models can be broadly categorized into two groups:

1. \textbf{Cascaded Approach}: These pipelines first apply an Optical Character Recognition (OCR) engine and then encode textual and visual features separately. Recent Arabic-focused examples include \emph{Arabic-Nougat}, which finetunes vision transformers to convert book pages into structured Markdown, handling multi-column layouts and diverse fonts \citep{rashad2024arabic}. Another example is \emph{Qalam}, a SwinV2-encoder + RoBERTa-decoder multi-modal LLM trained on over 4.5 million manuscript images, achieving under 1.2\% WER on printed Arabic and 0.8\% on handwriting \citep{bhatia2024qalam}.

2. \textbf{End-to-End Vision-Based Approach}: These models ingest raw document images and directly output text or structured representations, often via a unified transformer. Key Arabic and multilingual advances include \emph{GOT (OCR-2.0)} \citep{wei2024generalocrtheoryocr20}, a 580 M-parameter end-to-end model supporting slice- and whole-page inputs with long-context decoding. Another notable example is \emph{QARI-OCR}, which adapts Qwen2-VL to Arabic using massive synthetic data, achieving state-of-the-art CER 0.061 and robust layout handling \citep{wei2024general}.

Evaluations on \emph{KITAB-Bench} show that LVLMs (e.g., GPT-4o, Gemini-2.0-Flash, Qwen, AIN) outperform classic OCR by nearly 50\% in CER \citep{heakl2025kitab} yet still struggle with multi-page reasoning and cross-page dependencies. In other words, their ability to capture \emph{long document} phenomena, such as cross-page co-reference, evolving layouts, and dense interleaving of text, tables, and figures, remains under-explored. Robust evaluation on true long-context Arabic corpora is, therefore, a critical next step.

\subsection{Automated Data Annotation Systems}

Training LLMs or LVLMs at scale needs trillions of high-quality, well-annotated data points, which is out of human-alone annotation. Current  annotation systems tend to employ autonomous AI agents for synthesizing and validating labels with less human engagement.
\emph{LabelLerr's} pipeline manages self-correction and active-learning loops to label millions of images on its own, realizing a reduction in manual effort of over 50\% with accuracy over 90\% \citep{labellerr2024}. \emph{LandingAI’s} agentic document extraction uses vision-language agents to detect form fields, tables, and checkboxes and to generate structured schemas end-to-end, without human intervention \citep{landingai_agentic_document_extraction_2025}. In the Arabic domain, \emph{Arabic.AI’s} ecosystem enables template-driven report generation but still requires manual setup and is not tailored for raw document annotation tasks \citep{arabicai_agentic_ecosystem_2025}; likewise, \emph{UiPath’s} Active Learning DU pipeline incorporates human-in-the-loop guidance but offers limited support for right-to-left scripts and complex multi-column layouts \citep{uipath_active_learning_du_pipeline_2025}. To our knowledge, no such system fully automates long-context Arabic DU annotation, demonstrating the novelty and timeliness of our fully automated multi-LVLM interactive workflow.

\section{Fully Automated Workflow for Data Collection}

We constructed our long document Arabic corpus by automatic web crawling of a number of online repositories with a multi-stage filtering and normalization pipeline for breadth and fidelity. We initially discarded pages with fewer than the minimum characters, pages under restrictive licenses, and documents that are not suitable for automated QA generation. HTML content was extracted with a DOM clever scraper built on \emph{BeautifulSoup} \citep{richardson2007beautifulsoup}, and native PDFs were handled by \emph{pdfplumber} to maintain layout and pull out text blocks \citep{jsvine2020pdfplumber}. Scanned paper documents and images were read with \emph{Tesseract OCR} \citep{smith2007overview} with custom preprocessing (binarization, deskewing) to maximize legibility.

There were Arabic-specific problems that required additional steps. Right-to-left directionality and mixed Unicode encoding produced character misalignment, and we added a bidirectional-text handler based on the Unicode Bidirectional Algorithm \citep{unicode1996unicode}.

 With these unified preprocessing efforts, our dataset realizes multi-page coherence and varied layout coverage, laying a solid foundation for long document comprehension. In addition, the collected data spans across a variety of domains such as education, finance, governmental reports, news, social media, technical manuals, etc.

\begin{figure}[h!]
    \centering
    \includegraphics[width=1\linewidth]{  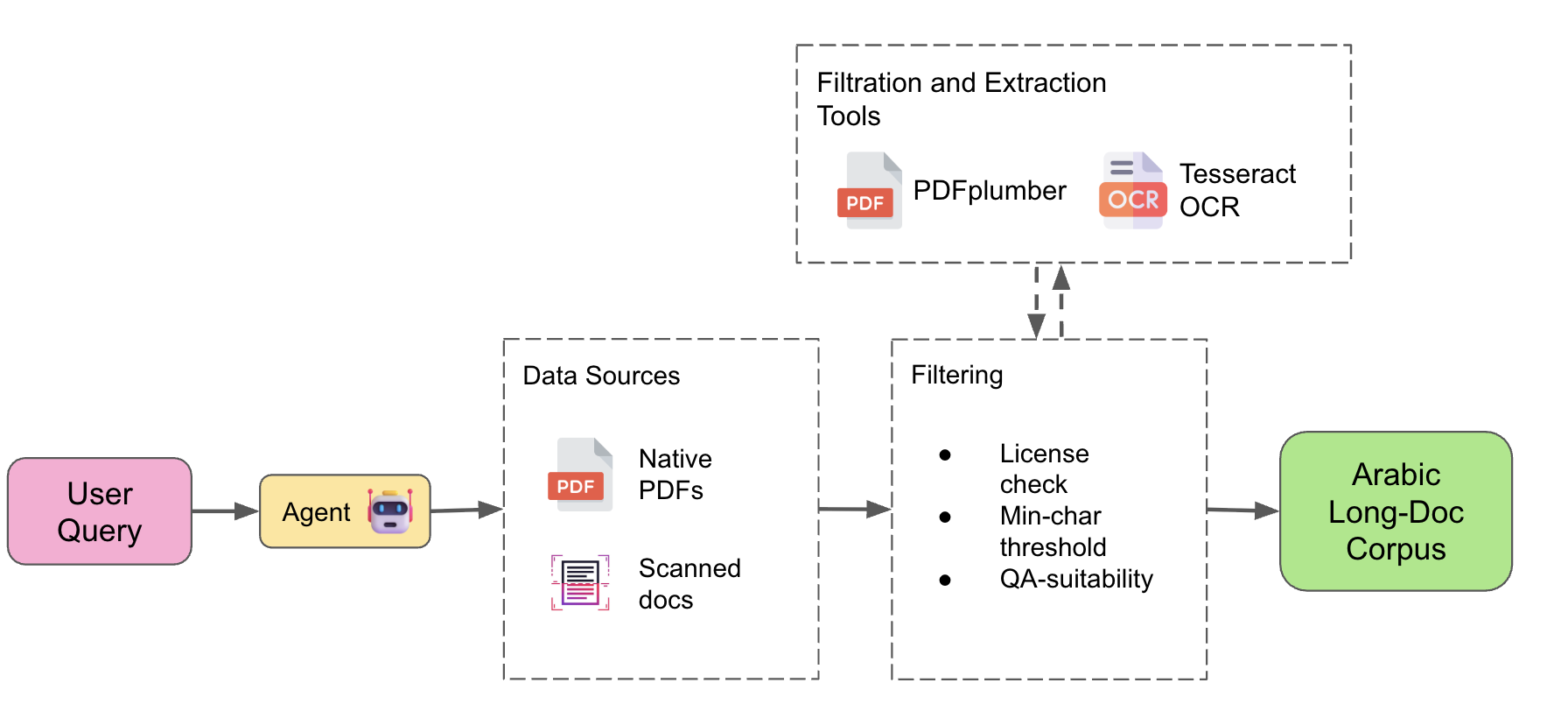}
    \caption{Automated Data Acquisition Workflow}
    \label{fig:workflow}
\end{figure}

Figure~\ref{fig:workflow} illustrates an end-to-end automated, LVLM-controlled process to build an Arabic long-document corpus.  The process involves four primary stages:

\begin{enumerate}[label=\arabic*.]
  \item \textbf{Agentic Query Dispatch:}
    For a high-level user request, an autonomous agent unit manages the following harvesting procedure, selecting appropriate repositories and search targets based on query semantics.
\item \textbf{Multi-modal Data Ingestion:}
    The agent retrieves candidate documents from diverse sources:
    \begin{itemize}
      \item \emph{Native PDFs:} Digitally created PDF documents downloaded from APIs or direct download.
      \item \emph{Scanned Documents:} Image-based documents (e.g., TIFF, JPEG) that require OCR to extract the text.
    \end{itemize}

  \item \textbf{Filtration and Extraction:}
    Raw inputs are processed by a modular toolset:
    \begin{itemize}
\item \texttt{PDFPlumber} to extract text and layout from native PDFs.
      \item \texttt{Tesseract OCR} to recognize scanned images as machine-readable text (accuracy is not a major concern at this stage) that enables character counts.
    \end{itemize}
    These components interact with the ingestion layer to support bidirectional refinement (e.g., re-crawling pages when layout anomalies are encountered).
  
  \item \textbf{Quality-Controlled Filtering:}
Automated screening of extracted documents is applied:
    \begin{itemize}
      \item \emph{License Compliance:} Checking against allowed reuse policies.
      \item \emph{Minimum Content Threshold:} Applying a character-count minimum to avoid evisceratingly brief texts.
      \item \emph{QA-Suitability Screening:} LVLM as a judge evaluation of each document's suitability for question-answer generation. To this end, we perform the filtering on a page-level with the document accepted as \say{QA-Suitable} only if $\geq 80\%$ of the pages pass the screening.
    \end{itemize}
Documents that satisfy all the criteria are aggregated into the final \emph{Arabic Long-Doc Corpus}, facilitating downstream tasks such as structured question generation and large-scale language modeling.
\end{enumerate}

\section{Self-Evolving  Adversarial QA Generator}
\subsection{Document Preprocessing}

The preprocessing phase transforms raw PDF inputs into structured representations suitable for downstream tasks, following these key steps:

\begin{itemize}[leftmargin=0.1in]
\item \textbf{PDF Ingestion:} The framework accepts documents in PDF format.
\item \textbf{Page Rasterization to Images ($I$):} PDF pages are converted into image format using \texttt{pdf2image} to maintain original visual layout and contextual details \citep{belval2018pdf2image}.
\item \textbf{Structural Layout Analysis ($L$):} A deep-learning model (e.g., \emph{DocLayout-YOLO}; \citep{zhao2024doclayout}) segments pages into logical elements such as headings, paragraphs, tables, and figures, enabling targeted content processing.
\item \textbf{Document Chunking with Overlap ($I_c$, $L_c$):}
In order to process long documents in an efficient manner, pages are segmented into overlapping chunks with length 50-page and 5-page overlap. It yields segmented images $I_c$ and structural layout annotations $L_c$ for each chunk.  Structural chunking was avoided due to the computational expense of page-level object detection and ordering, as well as the lack of availability of structural cues for scanned or poorly formatted documents. Fixed-size overlapping chunking was therefore selected for stability, scalability, and insensitivity to format variation.
\end{itemize}

\subsection{Self-Evolving  Adversarial Workflow}
\begin{figure}[h!]
    \centering
        \includegraphics[width=1\linewidth]{  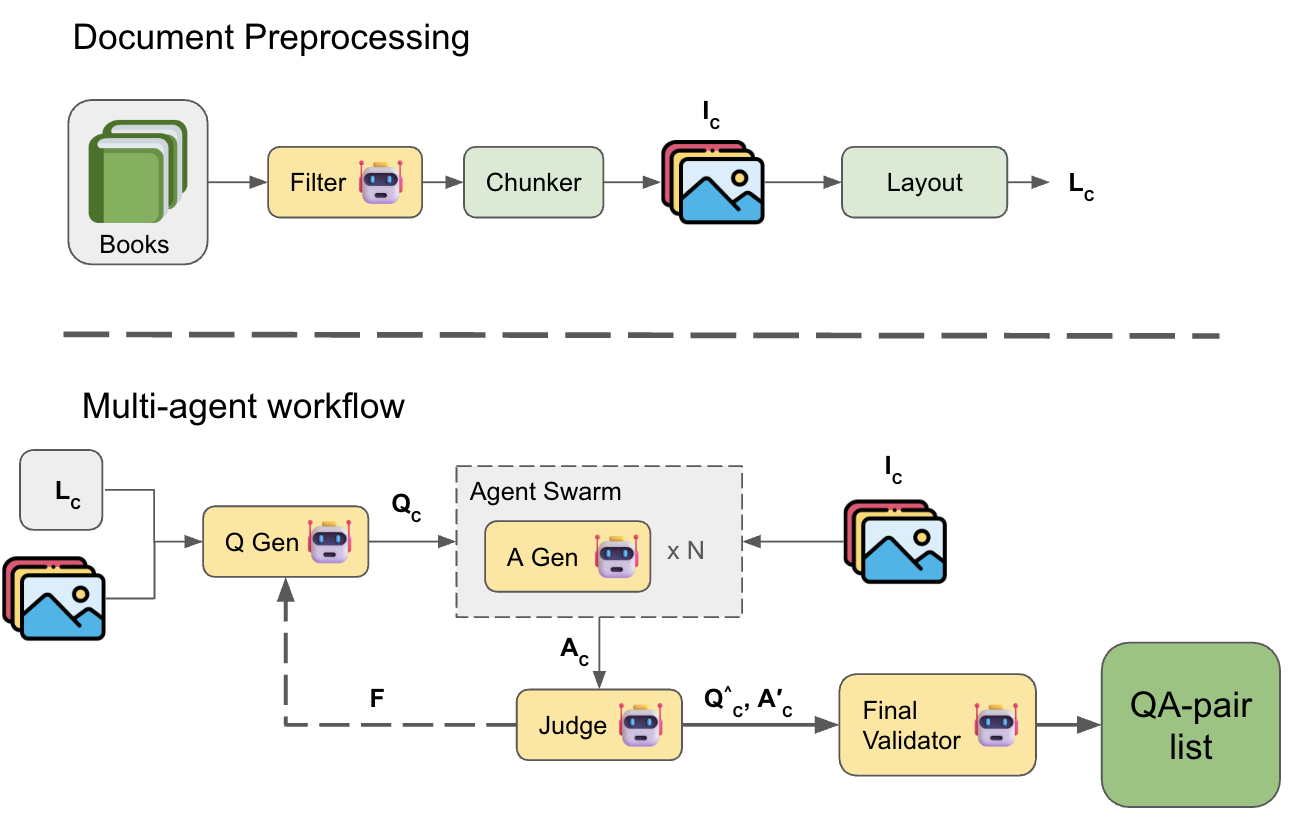}
    \caption{Self-evolving  Question Generation Workflow}
    \label{fig:QA}
\end{figure}
Following preprocessing, the multi-LVLM interactive workflow iteratively refines question-answer generation through the following structured sequence:

\textbf{Q Gen: Question Generation}
\begin{itemize}[leftmargin=0.1in]
\item \emph{Input:} Image(s) and layout annotations ($I_c$, $L_c$).
\item \emph{Output:} Draft question $Q_c$ (questions + cognitive premises), generated according to policy $\pi$, ensuring relevance and traceability to source content.
\end{itemize}

\textbf{Agent Swarm: Answer Generation}
    \begin{itemize}[leftmargin=0.1in]
        \item \emph{Input:} Image(s) $I_c$ and proposed question $Q_c$.
        \item \emph{Output:} Candidate answers $\{A_{c_i}\}_{i=1,\dots,N}$ (answers + logical foundations), where $N$ is the number of agents in the swarm, grounded explicitly in document content.
    \end{itemize}

\textbf{Judge: Assessment and Feedback}
    \begin{itemize}[leftmargin=0.1in]
        \item \emph{Input:} Full context data ($I_c, L_c$), questions $Q_c$, candidate answers $\{A_{c_i}\}_{i=1,\dots,N}$, and generation policy $\pi$.
        \item \emph{Output:} Validated answers $\{A^\prime_{c_i}\}_{i=1,\dots,N}$, question difficulty ratings, and actionable feedback $F$ (correct answer + attempted answer + evaluation + suggested refinement) for question improvement.
    \end{itemize}

\textbf{Q Gen: Iterative Question Refinement}
    \begin{itemize}[leftmargin=0.3in]
        \item \emph{Input:} Feedback $F$ from \textbf{Judge}.
        \item \emph{Output:} Further refined question $\hat{Q}_c$, iteratively cycling through Step 1 until the desired quality and consistency are achieved.
    \end{itemize}

\textbf{Final Validator: Evidence Validation}
    \begin{itemize}[leftmargin=0.3in]
        \item \emph{Input:} Comprehensive context data ($I_c, L_c$), proposed question $\hat{Q}_c$, and validated answers $A_c'$.
        \item \emph{Output:} Finalized questions $\hat{Q}_c$, each paired with rigorously validated evidence and answers.
    \end{itemize}

\textbf{Global Document Iteration:} the iterative loop described above is repeatedly executed on every segment of the document, establishing a complete, verified collection of question-answer pairs for the entire document.

This multi-LVLM and collaborative method through repeated refinement ensures contextual correctness and robust validation, making Long DU an effective instrument for large-scale and intricate document understanding tasks. Detailed illustrations of the workflow is documented in Figure~\ref{fig:QA}.

The workflow architecture utilizes prompted structured questions to sequence LVLM interaction on every step:
\begin{itemize}[leftmargin=.1in]
\item \textbf{Question Generation Prompt (QGP):} Telling \textbf{Q Gen} to create detailed and reflective questions at three levels of complexity:
\begin{itemize}[leftmargin=.1in]
\item \textbf{Level 1 (Factual):} Questions requesting explicit information extraction from the text.
\item \textbf{Level 2 (Inferential):} Questions requesting logical reasoning and inference based on contextual clues.
\item \textbf{Level 3 (Contextual Ambiguity):} Questions that are context-derived but explicitly unanswerable from the provided document.
\end{itemize}
\item \textbf{Question Refinement Prompt (QRP):} Guiding \textbf{Q Gen} to refine and improve the depth of its proposed questions based on the comprehensive feedback returned from \textbf{Judge}.
\item \textbf{Answer Generation Prompt (AGP):} Instructing the \textbf{Agent Swarm} to produce accurate, contextually appropriate, and well-supported answers.
\item \textbf{Assessment Prompt (AP):} Instructing \textbf{Judge} to evaluate question complexity, rejecting overly simplistic questions, and triggering iterative refinements towards improved quality.
\item \textbf{Evidence Validation Prompt (EVP):} Commanding \textbf{Final Validator} to validate the source (e.g., tables, text, charts, etc) of the answers returned from \textbf{Judge}.
\end{itemize}

\subsection{Iterative Refinement and Validation}

With repeated cycles of iterative multi-LVLM cooperation, questions persistently evolve to maximize coverage, depth, and relevance:
\begin{itemize}[leftmargin=.3in]
\item If \textbf{Judge} observes a greater than 50\% accuracy rate in some question, it notifies \textbf{Q Gen} to raise question complexity, thereby challenging the \textbf{Agent Swarm} to elevate the difficulty level of the proposed questions.
\item \textbf{Final Validator} strictly checks last question-answer pairs against verified sources, basically resolving contradictions and enhancing congruence against former observed benchmarks \citep{ma2024mmlongbench}.
\end{itemize}

\section{Data Analysis}
From the initial pool of 1,301 Arabic candidate documents, we have retained 113 after subjecting them to a multi-stage filtering pipeline with an ending acceptance rate of 8.6\%. The retained corpus spans a large number of domains including Legal (14), Medical (12), Research (10), Finance (10), Policy (9), Education (9), Manuals (8), News (8), Literature (8), Business (7), Technology (7), Environment (6), and History (5). Notably, OCR accuracy was not one of the most important issues in the recruitment process; OCR was employed solely as a surrogate to estimate character frequency and to verify that documents held a minimum of content.

The final dataset consists of well-structured tuples containing (question, answer, evidence pages, evidence sources, justification, and validation), making it a robust resource for long-document understanding research.

\begin{figure}[h]
    \centering
        \includegraphics[width=1\linewidth]{  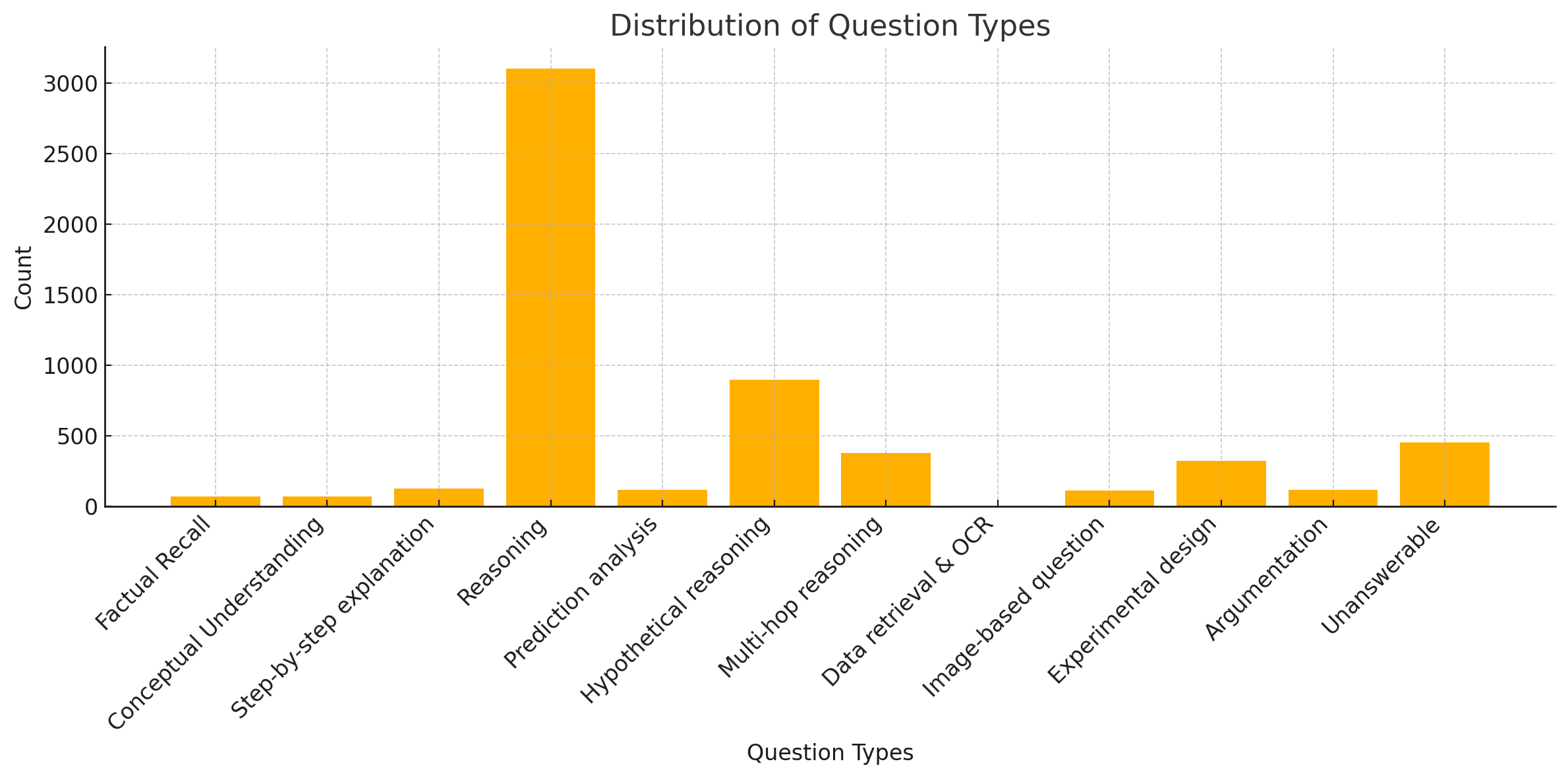}
    \caption{Proposed Question Types (50\% Accuracy Threshold)}
    \label{fig:pie_chart}
\end{figure}
Figure~\ref{fig:pie_chart} presents the histogram of 5,778 questions divided into twelve categories. Below is a detailed breakdown and interpretation.

\begin{itemize}[leftmargin=0.1in,label={}]
  \item \textbf{Core \& Hypothetical Reasoning (75.9\%)}  
    \begin{itemize}[leftmargin=0.3in]
      \item \textbf{Reasoning:} 3,103 (53.7\%)  
        Emphasizes logical deduction, inference, and problem-solving skills.
      \item \textbf{Hypothetical Reasoning:} 898 (15.5\%)  
        Probes “what-if” scenarios, testing flexible application of knowledge under counterfactual conditions.
      \item \textbf{Multi-hop Reasoning:} 381 (6.6\%)  
        Chains together multiple inference steps for deeper, integrative understanding.  
    \end{itemize}

  \item \textbf{Integrity Checks \& Multi-evidence Source Comprehension (9.9\%)}  
    \begin{itemize}[leftmargin=0.3in]
      \item \textbf{Unanswerable:} 454 (7.9\%)  
        Assesses the ability to withhold answers when no valid solution exists, reducing hallucinations.  
      \item \textbf{Image-based Question:} 112 (1.9\%)  
        Requires visual interpretation of charts, diagrams, or photographs.  
      \item \textbf{Data Retrieval \& OCR:} 6 (0.1\%)  
        Targets extraction of embedded or scanned text from documents.
    \end{itemize}

  \item \textbf{Intermediate-Complexity Tasks (9.6\%)}  
    \begin{itemize}[leftmargin=0.3in]

      \item \textbf{Experimental Design:} 321 (5.6\%)  
        Involves planning or critiquing scientific studies.  
      \item \textbf{Prediction Analysis:} 118 (2.0\%)  
        Requires forecasting outcomes based on provided data.  
      \item \textbf{Argumentation:} 116 (2.0\%)  
        Focuses on constructing or evaluating persuasive arguments.
    \end{itemize}

  \item \textbf{Basic Comprehension \& Procedural Explanation (4.6\%)}  
    \begin{itemize}[leftmargin=0.3in]
      \item \textbf{Step-by-step Explanation:} 126 (2.2\%)  
        Demands clear, ordered procedural breakdowns.  
      \item \textbf{Conceptual Understanding:} 72 (1.2\%)  
        Probes grasp of underlying principles rather than surface details.  
      \item \textbf{Factual Recall:} 71 (1.2\%)  
        Tests straightforward retrieval of explicit information.
    \end{itemize}
\end{itemize}
The dataset is heavily skewed toward reasoning (combined 75.9\%), which are typically the most challenging tasks that require intensive and profound thinking. The second most challenging group of tasks including integrity checks (unanswerable), multi-modal questions where the answers are based on numerous sources (e.g., tables, charts, images, etc), and OCR represent the second largest population (9.9\%) in the dataset. Tasks of intermediate complexity (9.6\%), covering multi-step inference, experimental planning, and argumentation, are the second largest population in the generated dataset.  Basic and simple comprehension and procedural items such as step-by-step explanation, conceptual understanding, and factual recall are the minority (4.6\%).

\section{Ablation Test}
In this section, we conduct an ablation test on the relationship between the accuracy threshold and the distribution of the proposed question types. In addition, we also verify that adding structural layout analysis increase the number of multi-modal questions. 
\begin{figure}[h!]
    \centering
        \includegraphics[width=1\linewidth]{  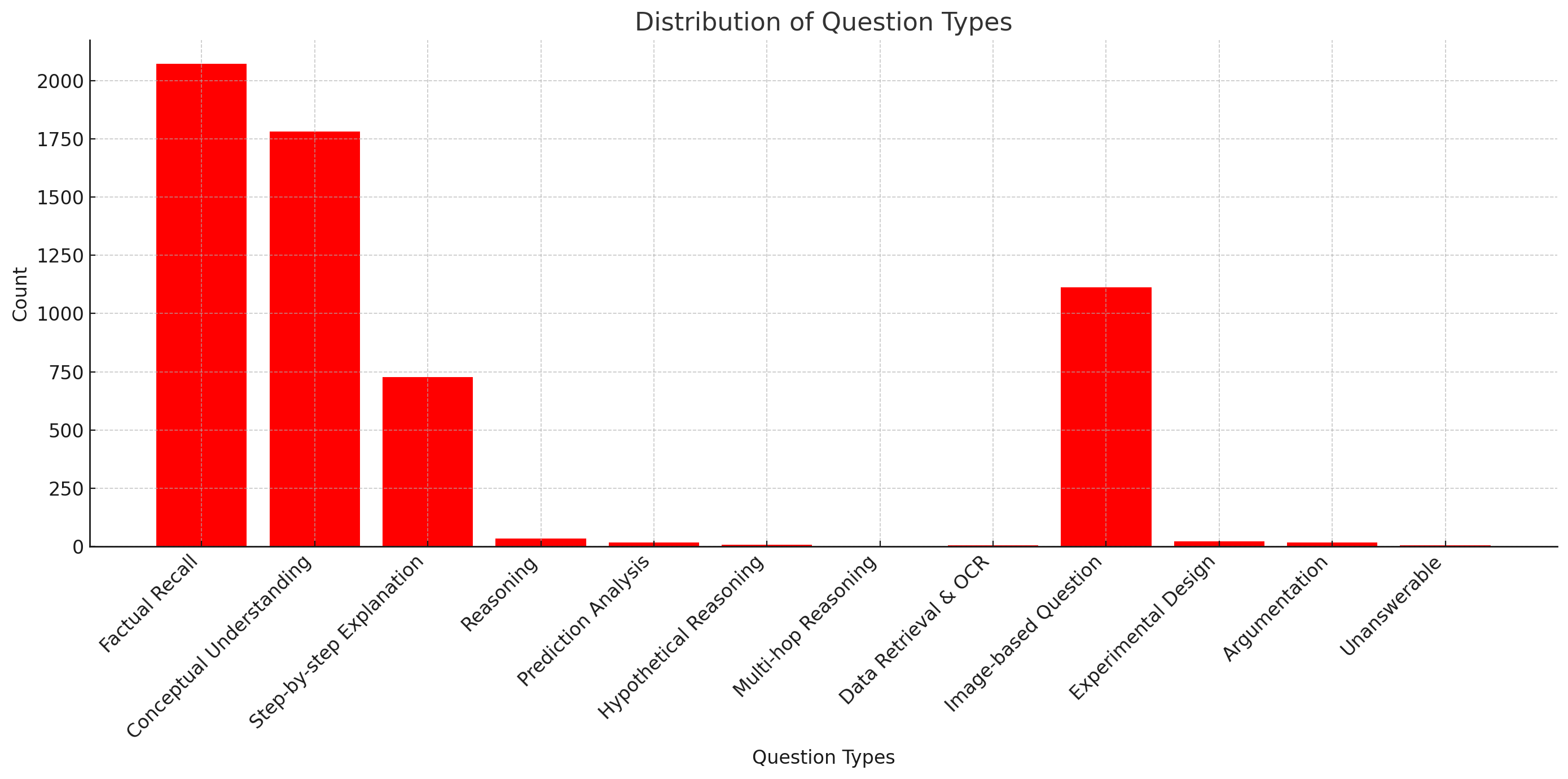}
    \caption{Proposed Question Types (No Accuracy Threshold)}
    \label{fig:pie_chart_2}
\end{figure}
\begin{figure}[h!]
    \centering
        \includegraphics[width=1\linewidth]{  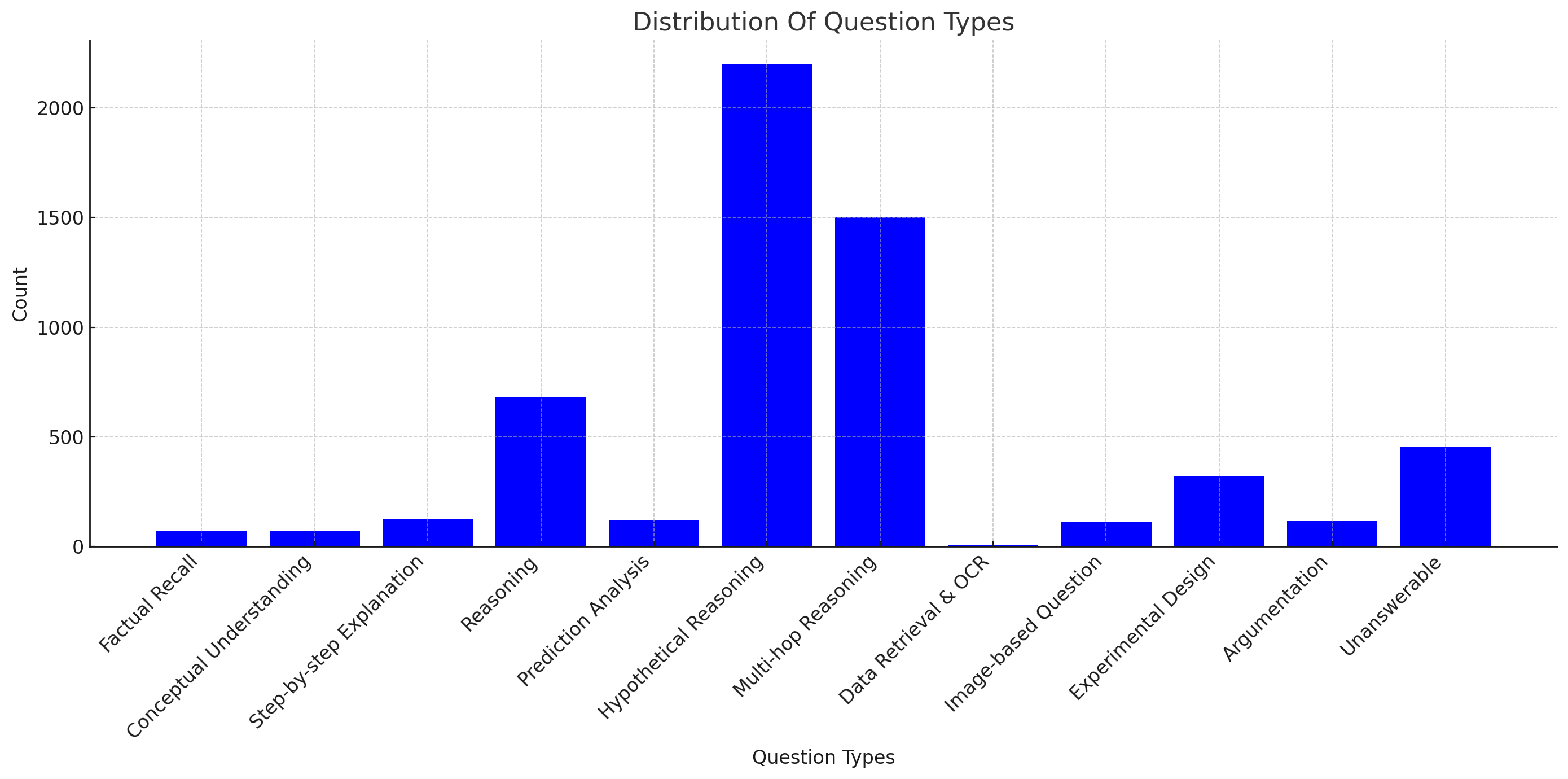}
    \caption{Proposed Question Types (25\% Accuracy Threshold)}
    \label{fig:pie_chart_3}
\end{figure}

By juxtaposing the original distribution (Figure~\ref{fig:pie_chart}), in which pure {\say{Reasoning}} questions dominated, with the low-threshold redistribution (Figure~\ref{fig:pie_chart_2}), we observe a striking re-balancing toward the hardest items. In fact, Figure~\ref{fig:pie_chart_3} reveals {\say{Hypothetical Reasoning}} swelling to around 38\% and {\say{Multi-hop Reasoning}} to around 26\%, with {\say{Reasoning}} itself now below 12\%.  At the same time, elementary tasks like factual recall and conceptual checks shrink to under 5\%, and mid-level formats (image questions, experimental design, argumentation) occupy only modest niches, while integrity checks (unanswerable/OCR) remain in place. In stark contrast, under extreme circumstances (without accuracy gate), Figure~\ref{fig:pie_chart_2} shows nearly two-thirds of questions as simple deductive inference and fewer than 15\% allocated to hypothetical or multi-hop chains. This side-by-side comparison confirms that relaxing accuracy constraints sharply redirects the generator’s output from straightforward inference toward the most complex, integrative reasoning challenges, ideal for stress-testing advanced models. 
\begin{table*}[h!]
    \centering
    \renewcommand{\arraystretch}{1.2} 
    \setlength{\tabcolsep}{4pt}      
    \resizebox{\textwidth}{!}{  
    \begin{tabular}{lcc|ccccc|ccccc|ccccc}
        \toprule
        \textbf{Model} & \textbf{Param} & \textbf{CW} 
        & \multicolumn{5}{c|}{\textbf{No Gate}}
        & \multicolumn{5}{c|}{\textbf{50\% Threshold}}
        & \multicolumn{5}{c}{\textbf{25\% Threshold}} \\
        \cmidrule(lr){4-8} \cmidrule(lr){9-13} \cmidrule(lr){14-18}
        & & & SC & MC & LC & SP & CP 
               & SC & MC & LC & SP & CP 
               & SC & MC & LC & SP & CP \\
        \midrule
        \multicolumn{18}{l}{\textbf{Closed-Source Models}} \\
        \midrule
        GPT-4o                   & --   & 128K 
                                 & 87.2\% & 84.5\% & 83.1\% & 90.9\% & 82.8\% 
                                 & 79.1\% & 78.2\% & 77.6\% & 83.5\% & 76.3\% 
                                 & 65.7\% & 61.5\% & 59.8\% & 71.2\% & 64.5\% \\
        Gemini-2.0 Flash         & --   & 1M   
                                 & \textbf{93.0\%} & 87.2\% & 85.5\% & \textbf{94.3\%} & 86.7\% 
                                 & \textbf{84.1\%} & 79.3\% & 79.0\% & \textbf{85.2\%} & {80.1\%} 
                                 & 71.8\% & 68.2\% & 67.5\% & 73.9\% & \textbf{72.0\%} \\
        Gemini-1.5 Pro           & --   & 2M   
                                 & 90.0\% & 86.3\% & 79.3\% & 91.2\% & 88.4\% 
                                 & 82.1\% & 79.0\% & 72.0\% & 83.0\% & 80.1\% 
                                 & \textbf{73.5\%} & {63.9\%} & 64.5\% & {68.3\%} & {67.6\%} \\
        Gemini-2.5 Pro           & --   & 2M   
                                 & 91.5\% & \textbf{89.0\%} & \textbf{88.2\%} & 93.4\% & \textbf{90.2\%} 
                                 & 81.7\% & \textbf{80.1\%} & \textbf{79.5\%} & 84.0\% & \textbf{81.3\%} 
                                 & {70.2}\% & \textbf{71.0\%} & \textbf{68.0\%} & \textbf{75.3\%} & 70.0\% \\
        \midrule
        \multicolumn{18}{l}{\textbf{Open-Source Models}} \\
        \midrule
        AIN                       & 7B   & 32K  
                                 & 78.5\% & 71.5\% & 67.7\% & 80.0\% & 71.5\% 
                                 & 69.1\% & 62.3\% & 60.2\% & 71.0\% & 62.0\% 
                                 & 58.2\% & 52.1\% & 49.3\% & 60.1\% & 50.6\% \\
        Aya Vision               & 32B  & 16K  
                                 & 79.1\% & 70.2\% & 66.7\% & 78.6\% & 70.4\% 
                                 & 68.7\% & 61.0\% & 58.0\% & 71.2\% & 60.2\% 
                                 & 57.3\% & 49.8\% & 46.9\% & 58.1\% & 50.0\% \\
        Qwen 2 VL                & 72B  & 32K  
                                 & 88.5\% & 84.0\% & {82.0\%} & 90.0\% & 83.5\% 
                                 & 78.7\% & \textbf{75.1\%} & \textbf{73.4\%} & 80.9\% & 74.0\% 
                                 & 68.4\% & 63.0\% & 61.2\% & 71.0\% & 63.9\% \\
        Qwen 2.5 VL              & 72B  & 128K 
                                 & \textbf{89.8\%} & \textbf{85.2\%} & \textbf{83.0\%} & \textbf{91.5\%} & \textbf{85.7\%} 
                                 & \textbf{79.6\%} & 74.8\% & 73.0\% & \textbf{82.0\%} & \textbf{74.5\%} 
                                 & \textbf{69.4\%} & \textbf{64.8\%} & \textbf{63.0\%} & \textbf{71.5\%} & \textbf{65.1\%} \\
        \bottomrule
    \end{tabular}
    }
    \caption{Combined performance of LVLMs on the \textbf{AraLongBench} across three accuracy conditions (No Gate, 50\%, 25\%) and varying context lengths and page conditions. CW = context window; SC = short-context; MC = medium-context; LC = long-context; SP = single-page; CP = cross-page.}
    \label{tab:model_arabic_combined}
\end{table*}
\begin{figure}[h!]
    \centering
        \includegraphics[width=1\linewidth]{  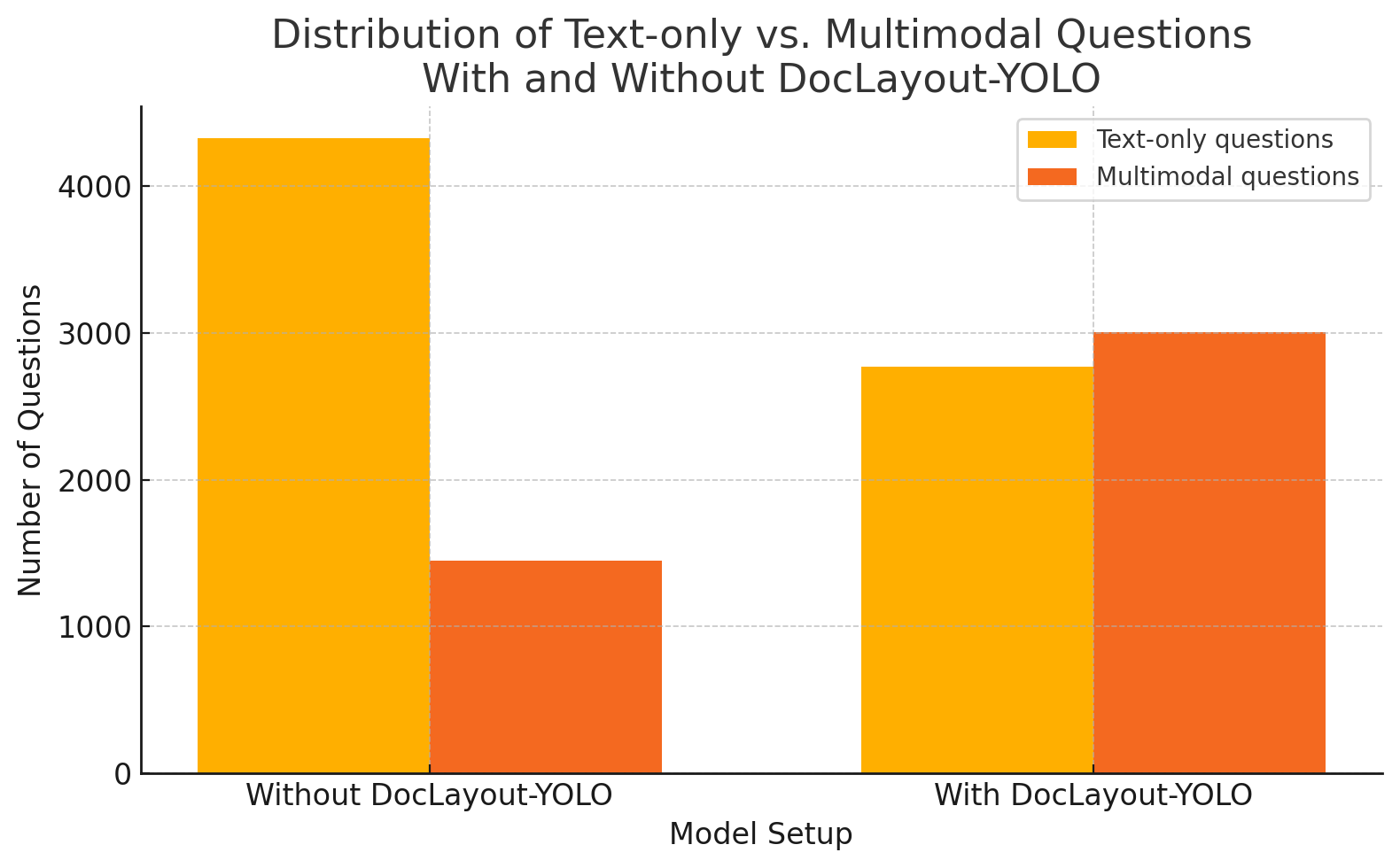}
    \caption{Distribution Of Text-Only Vs. Multi-modal Questions With and Without \emph{DocLayout-YOLO}}
    \label{fig:pie_chart_4}
\end{figure}

We apply \emph{DocLayout-YOLO} to conduct structural layout analysis and retrieve bounding boxes of  multi-modal elements such as tables, figures, charts, and other non-textual content. These regions are cropped and re-placed onto their corresponding document pages to form enriched page-level representations. Such a compositional strategy not only enhances multi-modal fidelity of information but also mitigates textual domination typically encountered in document processing pipelines, leading to a more semantically heterogeneous and balanced input space for downstream processing.

Figure~\ref{fig:pie_chart_4} illustrates how the introduction of \emph{DocLayout-YOLO} turns the rate of text-only questions against multi-modal questions.  Under the \say{Without \emph{DocLayout-YOLO}} mode, the system generated 4,327 text-only questions, nearly 75\% of the output, against 1,451 samples (25\%) that relied on visual content.  When \emph{DocLayout-YOLO} is introduced, however, the composition turns around: text-only questions fall to 2,771 (48\%), and multi-modal questions rise more than twice to 3,007 (52\%). This shift is something more than a statistical anomaly; it reflects a fundamental change in the preoccupation of the system.  By accurately detecting and leveraging document layout elements (tables, figures, diagrams), \emph{DocLayout-YOLO} unlocks a rich seam of visually grounded queries that were previously under-exploited.  The increased use of multi-modal items not only provides diversity and complexity to the item pool but also forces downstream models to have to join text and graphics, which matters a lot. The histogram shows that adding layout awareness created a shift in generator focus that first made the multi-modal question type less than 25 percent; now it is a majority.

Lastly, we also involved human efforts in verifying the practicality of the \textbf{Final Validator}. First of all, we removed the \textbf{Final Validator} from the workflow and generated data as usual. Then, we randomly sampled 100 questions to inspect the reported evidence sources and found a 14\% mismatch rate between the evidence source and the associated answer. The same process with this component back in the workflow was able to reduce the mismatch rate to below 5\%.

\section{Experiments}
Table~\ref{tab:model_arabic_combined} provides an extensive comparative analysis of the performance exhibited by open-source and closed-source LVLMs on a variety of accuracy levels on the newly developed Arabic benchmarks in this work with differing accuracy thresholds. The benchmarks test the models under a wide range of context settings, categorically distinguished as short-context (SC: $<100$ pages), medium-context (MC: $100$-$200$ pages), and long-context (LC: $>200$ pages), and tests based on single-page (SP) and cross-page (CP) conditions. Results are expressed as percentage accuracy to permit detailed observations on each model's performance in relation to the complexity of the task and linguistic intricacies inherent in Arabic.

Monotonically decreasing trends in performance are observed as we increase the accuracy bar across all models.  On \say{No Gate}, Gemini series and GPT-4o both get high-80s to low-90s across all page states and context lengths, demonstrating their true potential by being generously forgiving. A 50\% gate threshold eliminates predictions on the margin, and mean scores decrease by approximately 10-12 points (Gemini-2.0 Flash SC from 93.0\% to 84.1\%; GPT-4o SP from 90.9\% to 83.5\%). The most stringent 25\% gate again lowers performance, decreasing a further 10-12 points for Gemini-2.0 Flash SP from 94.3\% to 73.9\%, and GPT-4o SC from 87.2\% to 65.7\%. This progressive decline reflects the accuracy of each model decreasing as questions get harder and harder to generate.

The open-source set also demonstrates an equally robust sensitivity to threshold decrease.
AIN begins well at 78.5\% SC with no gate, drops to 69.1\% at 50\%, and further to 58.2\% at 25\%, a 20-point decline.  Aya Vision's decline is equally steep, dropping from 77.0\% to 68.7\% (50\%) and further to 57.3\% (25\%). Qwen 2 VL and Qwen 2.5 VL, although some of the strongest open models, follow this trend too: Qwen 2.5 VL's SP accuracy goes from 91.5\% (No Gate) to 82.0\% (50\%) and then from 71.5\% (25\%). Even top models lose around a 20-point difference under the toughest threshold. This cascading decline across open-source architectures reveals that with increasingly harder tasks, model confidence is lower.
\section{Limitations}
Although our self-learning Arabic QA system provides strong empirical gains and automaton benefits, its shortcomings remain. To begin with, as compelling as the system's performance on long-context Arabic documents is, its quality is highly sensitive to the structure and quality of input documents. High-visual-noise, scan-degraded, or non-standard layout documents, a common feature of historical Arabic collections, are capable of compromising the fidelity of the layout parser and impacting downstream QA accuracy.

Second, while fully automated workflow, current LVLM-based system relies on strict prompting templates and hard-coded complexity bounds as thresholds for validity checking and tuning.  Future updates would involve reinforcement learning or adaptive policy selection mechanisms in an effort to make more intelligent and adaptive prompting strategies.

Third, computational cost and system complexity are not to be underestimated. The multi-LVLM iterative pipeline, particularly in the self-refining stages, may be quite expensive in terms of latency and hardware. This can pose difficulties for real-time or mass deployment, especially where resources are limited.

Lastly, while we designed our architecture with Arabic-specific challenges in mind (e.g., bidirectional text, script variability), it remains to be seen how the system will perform across dialectal forms, handwriting material, or low-resource scripts generally within the broader Arabic linguistic context. Accommodating diverse regional Arabic dialects and mixed-script material remains an important area for further research.

\section{Conclusion}
In this work, we introduced a self-evolving  adversarial pipeline. Through the integration of state-of-the-art structural layout analysis and preprocessing, our pipeline not only successfully scales across diverse long-form texts but also strictly follows source content with fidelity in each iterative cycle.

One of the highlights of our system is its adaptive level of difficulty, enabling data generation from trivial recall drills to very advanced inference problems. The capability is present naturally for enabling curriculum-learning approaches, incrementing task difficulty over time in an effort to maximize model calibration and learning efficiency.

The methodology will be applied to other domains and modalities in the future, be computationally optimized, and be introduced with adaptive thresholding for adaptive data generation.

In addition, we have developed an automated data-acquisition pipeline that transforms a single, high-level user query into a fully automated, multi-stage harvesting process capable of gathering millions of documents in a matter of hours with an easy-to-use interface and end-to-end automation that eliminates manual data-collection bottlenecks. Because of its agent-based, modular design, the pipeline is readily extensible to new domains and languages far beyond the scope of Arabic document understanding such as legal rulings, medical literature reviews, or multilingual scientific benchmarks.

\bibliography{custom}

\end{document}